\documentclass[]{article}
\usepackage[a4paper]{geometry}
\usepackage{mtsummit2023}
\usepackage{times}
\usepackage{url}
\usepackage{latexsym}
\usepackage{natbib}
\usepackage{layout}
\usepackage[hang,marginal]{footmisc}
\usepackage{graphicx}
\usepackage{multirow}
\usepackage{subcaption}
\usepackage{arydshln}
\usepackage{xcolor}
\usepackage{hhline}
\usepackage{booktabs}

\begin{document}

\title{Perturbation-based QE: An Explainable, Unsupervised Word-level Quality Estimation Method for Blackbox Machine Translation}

\author{\name{\bf Tu Anh Dinh} \hfill  \addr{tu.dinh@kit.edu}\\
        \addr{Karlsruhe Institute of Technology, Germany}
\AND
        \name{\bf Jan Niehues} \hfill \addr{jan.niehues@kit.edu}\\
        \addr{Karlsruhe Institute of Technology, Germany}
}

\maketitle
\pagestyle{empty}

\begin{abstract}
Quality Estimation (QE) is the task of predicting the quality of Machine Translation (MT) system output, without using any gold-standard translation references. 
State-of-the-art QE models are supervised: they require human-labeled quality of some MT system output on some datasets for training, making them domain-dependent and MT-system-dependent. 
There has been research on unsupervised QE, which requires glass-box access to the MT systems, or parallel MT data to generate synthetic errors for training QE models. 
In this paper, we present \textit{Perturbation-based QE} - a word-level Quality Estimation approach that works simply by analyzing MT system output on perturbed input source sentences.
Our approach is unsupervised, explainable, and can evaluate any type of blackbox MT systems, including the currently prominent large language models (LLMs) with opaque internal processes. 
For language directions with no labeled QE data, our approach has similar or better performance than the zero-shot supervised approach on the WMT21 shared task. 
Our approach is better at detecting gender bias and word-sense-disambiguation errors in translation than supervised QE, indicating its robustness to out-of-domain usage. The performance gap is larger when detecting errors on a nontraditional translation-prompting LLM, indicating that our approach is more generalizable to different MT systems.
We give examples demonstrating our approach's explainability power, where it shows which input source words have influence on a certain MT output word.
\end{abstract}

\section{Introduction}
Machine Translation (MT), with the aim of translating text from a source language to a target language, has been increasingly adopted in different real-world scenarios, ranging from translations in healthcare areas to translations in the legal domains \citep{vieira2021understanding}. 
In many of these applications, errors in translation output could cause serious harm to the users, e.g., translation errors leading to wrong medical diagnoses in healthcare or wrong judgment in court. Therefore, it is important to let the users know how much they can trust a translation, by providing them with some quality assessment of the MT output. 
This is not always straightforward due to the lack of gold-standard human translations, or the mismatch between evaluation data and real-world usage.
As a result, researchers have been looking into Quality Estimation. 

Quality Estimation (QE) is the task of predicting the quality of MT system output without access to reference translations. 
State-of-the-art QE systems are built in a supervised manner, where they require human-labeled quality assessment on MT output for training \citep{rei2022cometkiwi}. 
This approach has 2 drawbacks: the labeled QE data is costly to obtain, and the trained QE models would only know about the types of error that are presented in the training data. Supervised QE models are likely to underperform in unfamiliar settings \citep{kocyigit-etal-2022-better}, e.g., when evaluating the output of a new MT system on a new dataset from a different domain.
Consequently, there has been research into unsupervised QE, where the human-labeled assessment data is no longer required \citep{fomicheva-etal-2020-unsupervised, tuan-etal-2021-quality}. 
These works either require glass-box information of the MT system (e.g., output log probabilities or attention scores), or a large amount of parallel MT data to create synthetic QE data for training. 
This is problematic for language pairs with low-resourced MT data, or when the MT system is kept blackbox, which is the current trend of some widely-discussed API-only large language models. 

In this paper, we propose an \textbf{unsupervised} word-level QE approach to evaluate \textbf{blackbox} MT systems, termed Perturbation-based QE. 
Our motivation is inspired by a known problem: when uncertain, MT systems rely on spurious correlations learnt from the training data to generate translation \citep{emelin-etal-2020-detecting, savoldi2021gender}. We assume that, when outputting a translation token, if the MT system relies on too many parts of the source sentence, it is likely that the system is exploiting irrelevant correlations, thus the output token is unreliable. 
Consider the English $\rightarrow$ German example: \textit{``My friend has a Ph.D. degree, and now she is a professor."} $\rightarrow$ \textit{``Meine Freundin hat einen Doktortitel, und sie ist jetzt eine Professorin."}. 
The translation word \textit{``Freundin"} should only depend on \textit{``friend"} and \textit{``she"}, where \textit{``friend"} indicates the meaning and \textit{``she"} indicates the gender form.
The output word being influenced by more source words would indicate that the MT system is focusing on the wrong part of the input sentence.

Broadly speaking, in Perturbation-based QE, we perturb words in the source sentences one by one to find out which source words influence a single output word. 
If an output word is influenced by too many source words, then it is predicted as a bad translation. 
Due to its simplicity, our approach does not require human-labeled QE data, nor parallel MT data, nor glass-box access to the evaluated MT system. 
Additionally, our QE approach comes with explainability power: it shows which source words affect each output word in the translation, thus can be used as an indication of the wrong correlations that is inherent in the MT system.

To summarize, our contributions are as follows:
\begin{itemize}
    \item Proposing Perturbation-based QE\footnote{Implementation available at \url{https://github.com/TuAnh23/Perturbation-basedQE}.}: a simple word-level Quality Estimation approach that is explainable, unsupervised and works with any type of blackbox MT systems, including the API-only large language models (i.e., MT-system-agnostic). 
    \item Experiments showing the advantages of Perturbation-based QE: (1) it has similar or better performance than zero-shot QE, without making use of labeled data of auxiliary language pairs 
    and (2) it is domain-independent and MT-system-independent compared to supervised QE methods: it can better capture out-of-domain gender errors and word-sense-disambiguation errors, especially from an unseen, nontraditional translation-prompting large language model and (3) it is not sensitive to hyperparameters.
    \item Analysis showing an example use of the explainability power of Perturbation-based QE.
\end{itemize}

\section{Related work}
Quality Estimation (QE) aims to predict the quality of Machine Translation (MT) output, either at the sentence level or word level. 
For word-level QE, the goal is to predict whether each word in the translation is correct. 
State-of-the-art word-level QE methods are supervised \citep{kim2017predictor, specia-etal-2021-findings}, i.e., requiring labeled data for training, which is costly to obtain. 
Additionally, supervised QE is likely to be domain-dependent and MT-system-dependent, as they do not aware of errors not occurring in the training data \citep{kocyigit-etal-2022-better}. 

Unsupervised QE overcomes the need for labeled data. 
Several works perform unsupervised QE by using glass-box features from the MT systems \citep{popovic-2012-morpheme, moreau-vogel-2012-quality, etchegoyhen-etal-2018-supervised, niehues-pham-2019-modeling, fomicheva-etal-2020-unsupervised}. As an example, \cite{fomicheva-etal-2020-unsupervised} proposed unsupervised QE using the output probability distribution and the attention mechanism from encoder-decoder MT models. Therefore, their methods are model-specific. 
\cite{tuan-etal-2021-quality} excludes the need for human-labeled data and MT glass-box access by creating synthetic data to train QE models. 
The synthetic data is generated by aligning candidate MT translations to the target references to find errors, or rewriting target reference sentences using a masked language model to introduce errors. 
These methods require a large and diverse amount of parallel MT data (i.e., source sentences and gold-standard translations), which is not always available for different domains and language pairs. 
Additionally, these methods are also likely to be domain-dependent and MT-system-dependent, as the QE model is trained on the output of pre-selected MT systems on pre-selected MT data.

Researchers also focus on Quality Estimation from the explainability perspective. 
\cite{he-etal-2019-towards} propose using integrated gradients from MT models (i.e., glass-box information) to quantify how important each source word is to the output translated words. 
The method is then used for QE by detecting under-translated source words that have low importance to the output translation. 
\cite{ferrando-etal-2022-towards} also quantify the contribution of each source word on the output translation using glass-box information from transformer-based MT models, which is the layer-wise tokens attributions. 
Here the source words' contribution can also be used to detect under-translated source words, or to assess the quality of the whole translation. 
Another line of research is on explainable sentence-level QE, where the word-level error scores are provided as the explanation for the predicted sentence-level score \citep{fomicheva-etal-2021-eval4nlp}. 
Explanations can be extracted by using methods such as LIME \citep{ribeiro2016should} or SHAP \citep{lundberg2017unified} on top of sentence-level QE models, or building interpretable models that output both sentence-level quality and word-level explanations \citep{fomicheva-etal-2021-eval4nlp}.

In contrast to the previous works on Quality Estimation, Perturbation-based QE does not require labeled QE data, parallel MT data, nor glass-box access to the evaluated MT system.
From the explainability perspective, our approach provides a new type of explanation for target-side word-level QE, i.e., the information on which source words affect each translated word. 

\section{Perturbation-based Quality Estimation} \label{sec:method}
\begin{figure*}[!ht]
  \centering
  \includegraphics[width=0.73\textwidth]{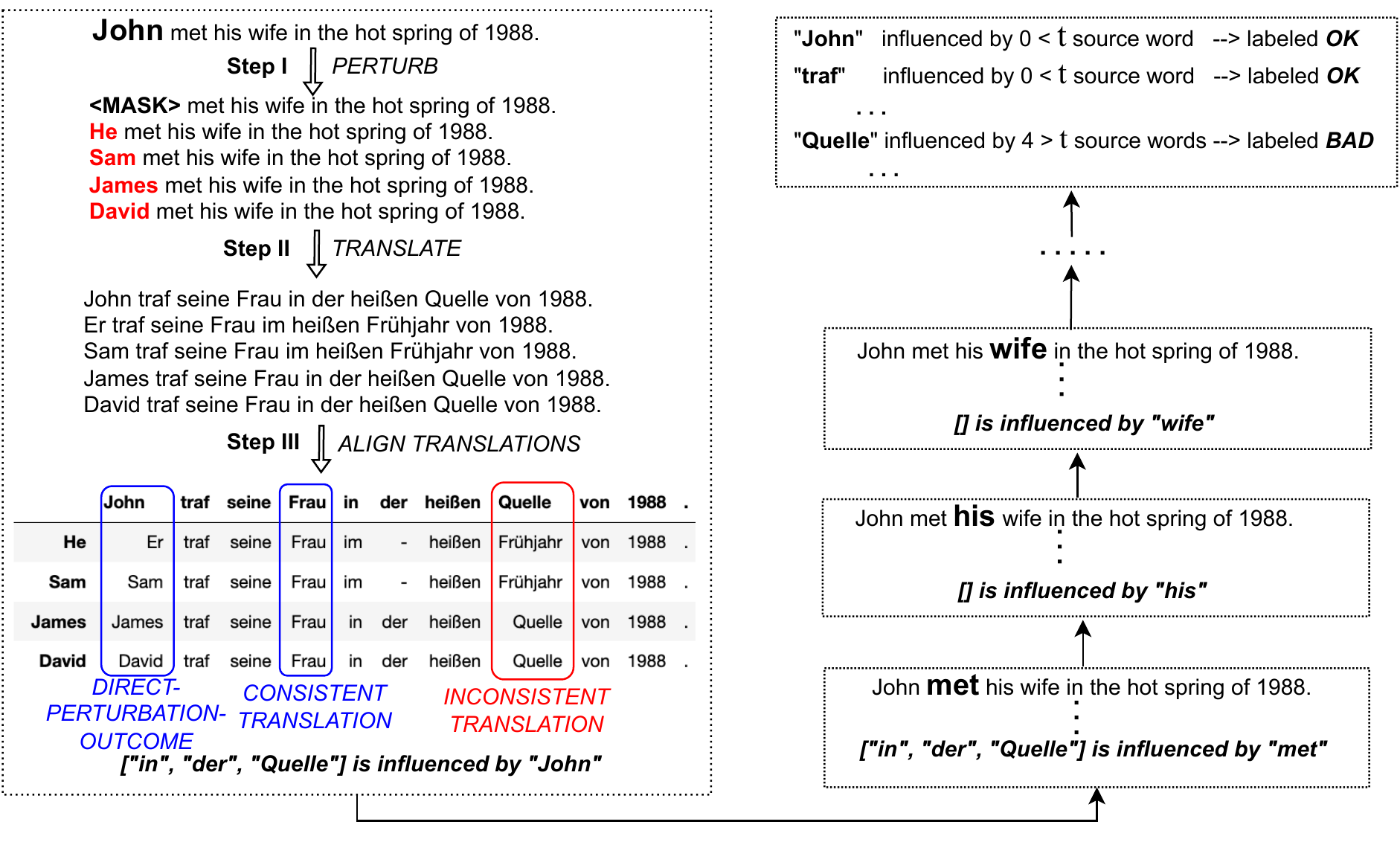}
  \caption{Perturbation-based QE. Words in the source sentence are perturbed one by one to find out their influence on the output words. If an output word $h_{j}$ is influenced by more than $t$ source words (excluding the source word directly translated to $h_{j}$), it is predicted as a BAD translation.}
  \label{fig:PerturbationBasedQE}
\end{figure*}
In this section, we describe Perturbation-based QE. 
Recall our motivation: if the MT system relies on too many tokens in the source sentence to output a translation token, it is likely that the system is exploiting irrelevant correlations, thus the translation token is unreliable.


\textbf{Perturbation generation} (Step I Figure \ref{fig:PerturbationBasedQE}): We first perform perturbation to the source sentence. 
The subset of source words to perturb, which is a hyperparameter choice, is one of the following: (1) \underline{content words}, i.e., noun, verb, adjective, adverb, pronoun, determined by NLTK part-of-speech tagging \citep{bird2009natural}; (2) \underline{all words} including functional words such as ``a", ``an", ``the"; or (3) \underline{all tokens} including non-word tokens such as punctuation marks. 
For each perturbed source word $s_i$, we mask it out from the source sentence and use a language model to generate $n$ best replacements. The language masking model can be BERT \citep{BERT}, RoBERTa \citep{liu2019roberta} or DistilBERT \citep{sanh2019distilbert} (choice of the language masking model is a hyperparameter).

\textbf{Translation} (Step II Figure \ref{fig:PerturbationBasedQE}): We use the MT system to translate all perturbed versions. 

\textbf{Alignment} (Step III Figure \ref{fig:PerturbationBasedQE}): We align at word level all the perturbed translations with the original translation. Two possible alignment methods are (1) Levenshtein \citep{levenshtein1966binary}, which is the standard edit-distance alignment method that minimizes the number of insertion, deletion and substitution operations; and (2) Tercom \citep{snover2006study}, which additionally considers the shift operation. 
In Figure \ref{fig:PerturbationBasedQE}, the alignment outcome is shown in a table, where the column titles are the tokenized original translation, the row titles are the replacements of the perturbed source word, and each row is the aligned translation of the perturbed source sentence. Note that sometimes the alignments are not one-to-one. Some words in the original translation could have no aligned version in the perturbed translation. In this case, we align the original word with an empty token. Similarly, some words in the perturbed translation might not be aligned with any word in the original translation. In this case, we discard the words in the perturbed translation, since we only evaluate the consistency of the original words.

\textbf{Consistency evaluation} (Step III Figure \ref{fig:PerturbationBasedQE}): An MT-output word $h_{j}$ is considered either a consistent translation, an inconsistent translation, or a direct-perturbation-outcome w.r.t. each perturbed source word $s_{i}$.
\begin{itemize}
    \item Consistent translation is a translation that remains the same across perturbations. For example, in Figure \ref{fig:PerturbationBasedQE}, \textit{``Frau"} is a consistent translation w.r.t perturbing \textit{``John"}. To account for possible noise in alignment, we mark a translation as consistent if it remains the same across more than $c\%$ out of $n$ perturbations w.r.t $s_{i}$. The threshold $c$ is a hyperparameter.
    \item Direct-perturbation-outcome is the translation of the perturbed word, thus should vary in all perturbations. For example, in Figure \ref{fig:PerturbationBasedQE}, \textit{``John"} is a direct-perturbation-outcome translation w.r.t perturbing \textit{``John"}. To account for possible noise in translation and alignment, we mark a translation as direct-perturbation-outcome if the number of unique versions over the total $n$ perturbations is larger than $p\%$. The threshold $p$ is a hyperparameter.
    \item Inconsistent translation is a translation that has a few versions of it across $n$ perturbations (i.e., the remaining cases).  In Figure \ref{fig:PerturbationBasedQE}, \textit{``Quelle" }is an inconsistent translation w.r.t perturbing \textit{``John"}. When an MT-output word $h_{j}$ is inconsistent due to perturbing a source word $s_{i}$, we say that $h_{j}$ is influenced by $s_{i}$. Here \textit{``Quelle" } is influenced by \textit{``John"}.
\end{itemize}

\textbf{Quality label prediction} (Last block Figure \ref{fig:PerturbationBasedQE}): If the number of source words influencing $h_{j}$ (excluding the one directly translated to $h_{j}$) is higher than a threshold $t$, then $h_{j}$ is predicted as a BAD translation, otherwise predicted as OK\footnote{
We focus on evaluating the translation words that were output by the MT system. Our approach is not suitable to evaluate the gap between words or to detect untranslated parts of the source sentence.}. 
The threshold $t$ is a hyperparameter.

Our approach comes with several advantages. First, it is unsupervised. The method does not rely on any labeled QE data or parallel MT data for training.
This potentially makes the approach domain-independent and MT-system-independent. In other words, the approach would be robust to discover errors not presented in previous datasets, such as errors from a new MT system on a different domain.
A small amount of labeled QE data can be used for hyperparameter tuning.
However, our experiments show that the approach is not sensitive to hyperparameter choices, and that hyperparameters can be transferred across languages.
Second, our approach is MT-system-agnostic and works for blackbox MT systems, as it only uses the MT system to generate translations.
Third, our approach comes with explainability power. For each MT output word, our method shows which source words affect the generation of the considered output word. In this way, one can find wrong correlations inherent in the MT systems. 

In terms of computational cost, Perturbation-based QE does not involve any training process. However, it requires computational power when using the evaluated MT system to generate translations of different perturbed versions of the source sentence. This can be considered as the trade-off between our approach and the previous QE approaches.

\section{Experimental setup}
\subsection{Overall evaluation} \label{sec:Overall_evaluation}

\textbf{Dataset}: We use the word-level part of the MLQE-PE dataset \citep{fomicheva2020mlqe}, which is the benchmark in the WMT21 QE shared task \citep{specia-etal-2021-findings}. The dataset consists of source sentences, the machine translation output and the word-level \textit{OK/BAD} labels. 
We conduct experiments on four language pairs: English-German (\textit{en-de}), English-Chinese(\textit{en-zh}), English-Japanese (\textit{en-ja}) and English-Czech (\textit{en-cs}). 
In this dataset, \textit{en-de} and \textit{en-zh} directions are supervised, while \textit{en-ja} and \textit{en-cs} directions are zero-shot. However, we only use the development split for \textit{en-de} and \textit{en-zh} to perform hyperparameter tuning. 

\textbf{Evaluated MT systems}: We use the to-be-evaluated encoder-decoder MT systems from WMT21, i.e., the fairseq Transformer \citep{ott-etal-2019-fairseq} bilingual models for \textit{en-de} and \textit{en-zh}; and the ML50 fairseq multilingual Transformer model \citep{tang2020multilingual} for \textit{en-cs} and \textit{en-ja}. 

\textbf{Metrics}: Following the WMT21 shared task, we use the Matthews correlation coefficient (MCC) \citep{matthews1975comparison} as the evaluation metric for word-level QE in our experiments.

\textbf{Hyperparameters}: Hyperparameters for our approach, as explained in Section \ref{sec:method}, are the number of unmasking replacements $n$, thresholds $c$, $p$, $t$, choices of source word subset for perturbation, language masking models to generate perturbation replacements and alignment tools. 
We use grid search to find the hyperparameter setting that yields the highest MCC score on development data. 
The best setting for \textit{en-de} (which is then applied on \textit{en-cs}) is $n=30$, $c=0.95$, $p=0.9$, $t=2$, perturbing content words, Tercom alignment and RoBERTa unmasking. 
The best setting for \textit{en-zh} (which is then applied on \textit{en-ja}) is $n=30$, $c=0.95$, $p=0.8$, $t=4$, perturbing all tokens, Tercom alignment and RoBERTa unmasking. 
We transfer the hyperparameters across languages in such a way since we expect more language-similarity between \textit{de/cs} (alphabetic writing systems) and \textit{zh/ja} (logographic writing systems). 

\textbf{Unsupervised QE baselines}: For unsupervised baseline, we use the word-level log probabilities generated by the MT system. If the log probability of an output word is larger than a certain threshold $l$, then it is marked as OK, otherwise it is marked as BAD. The threshold $l$ is a hyperparameter. Here we also use the development split to find the best $l$ for \textit{en-de} and \textit{en-zh}. We apply the best value of $l$ for \textit{en-de} (which is $log_2 0.45$) on \textit{en-cs} and the best value of $l$ for \textit{en-zh} (which is $log_2 0.60$) on \textit{en-ja}. 
We choose this baseline since it has the same data usage as our approach, and it requires little information from the MT system (although here we no longer treat the MT system completely as blackbox). 

\textbf{Supervised QE baselines}: We use the supervised baseline from the WMT21 QE shared task \citep{specia2021findings}. The baseline is a multilingual transformer-based Predictor-Estimator \citep{kim-etal-2017-predictor}, trained on labeled data for all available seven language directions.
The model is trained multi-tasked, requiring both word-level and sentence-level labeled data.  

\subsection{Out-of-domain, unseen-MT-system evaluation}
\textbf{Common in-domain, known-MT-system setup}: The common evaluation setup for QE approaches, e.g., in the WMT21 shared task, are in-domain and on known MT systems. That is, the QE test data is generated in the same way as the QE training data, and the to-be-evaluated MT system is the same as the one used to create the QE training data. 
However, in order to be useful in real-world applications, QE approaches should be capable of out-of-domain evaluation on unseen MT systems. 
That is, QE approaches should be able to evaluate different types of MT systems on different types of datasets. Therefore, we design experiments using QE approaches in an out-of-domain, unknown-MT-system setting, described as follows.

\textbf{Evaluated MT systems}: We test the QE approaches on evaluating two MT systems, one known and one unseen. The known MT system is the one that was used to create the WMT21 QE training and test data: the Fairseq encoder-decoder MT model. The unseen system is Flan-UL2 (available on HuggingFace) - a recent prompt-based large language model (LLM). We generate MT output from this LLM by prompting the system with ``\textit{Translate this into German:} $<$\textit{English\_input}$>$.". We choose this system as LLMs have been gaining a lot of attention and are more and more widely used \citep{vilar2022prompting, zhang2023prompting, bawden2023investigating}. Going beyond the conventional encoder-decoder MT systems, we attempt to show that our approach is applicable to prompt-based translation using these prominent decoder-only LLMs.

\textbf{Out-of-domain test data}: We use two challenge sets on \textit{en-de}. The first one is WinoMT \citep{stanovsky-etal-2019-evaluating}, used to evaluate gender bias from MT systems. 
WinoMT contains English input sentences with marked gender roles (e.g., “The \textbf{doctor} asked the nurse to help her in the operation”) and evaluation protocol to identify whether the MT system outputs the correct gender form. 
The second challenge set is MuCoW (WMT 2019 translation test suite version) \citep{raganato-etal-2019-mucow}, used to evaluate word-sense-disambiguation ability of MT systems. MuCoW contains English input sentences with ambiguous words and evaluation protocol to identify whether the MT system outputs the correct sense translations of the ambiguous words. 

On WinoMT, the correct-gender accuracy is 69.4\% for the Fairseq encoder-decoder MT system and 47.5\% for the Prompt-based LLM Flan-UL2 system. 
On MuCoW, the correct-disambiguation accuracy is 47.59\% for the Fairseq encoder-decoder MT system and 22.95\% for the Prompt-based LLM Flan-UL2 system.
Both MT systems do not perform well in outputting the correct gender form nor outputting the correct sense for ambiguous words.
Therefore, it would be interesting to see whether QE methods can detect these mistakes.


\textbf{Out-of-domain error detection}: We test whether QE approaches can detect gender errors (which we refer to as GenderBAD tokens) and word-sense-disambiguation errors (which we refer to as WSD-BAD tokens). Given an MT system, we first generate translations for the WinoMT/MuCoW English sentences. Then we run WinoMT/MuCoW evaluation protocol to mark the GenderBAD/WSD-BAD tokens. An ideal QE approach should be able to detect all the GenderBAD/WSD-BAD tokens, i.e., correctly labeling them as BAD translations. 

\textbf{Metrics}: We report on the GenderBAD-recall and WSD-BAD-recall, i.e., the percentage of GenderBAD/WSD-BAD tokens (marked by WinoMT/MuCoW) that are successfully predicted as BAD by the QE methods. 
We do not report on the GenderBAD/WSD-BAD accuracy, since tokens with correct gender form or correct disambiguated sense are not necessarily OK translations. They could contain some other types of errors such as tense or singular/plural forms.
Note that the recall metric could favor QE methods that are overly harsh (e.g., predicting everything as BAD would result in perfect GenderBAD/WSD-BAD recall). 
Therefore, we additionally report on the GenderBAD-precision and WSD-BAD-precision. Precision scores only serve as an indication of whether a QE model is too harsh. 
The reason is that, GenderBAD and WSD-BAD are not the only types of BAD error, thus it is not correct to always punish the QE model for predicting a non-GenderBAD or non-WSD-BAD token as BAD.


\subsection{Robustness w.r.t hyperparameter choices} \label{sec:robust_hyp_exp}
Recall that, in the previous experiments, we use the WMT21 development data of \textit{en-de} and \textit{en-zh} to perform hyperparameter tuning for Perturbation-based QE. The best hyperparameters for \textit{en-de} are then also used for \textit{en-cs}, and the ones for \textit{en-zh} are used for \textit{en-ja}, since we assumed more language similarity between \textit{de/cs} and \textit{zh/ja}. We refer to this as \textit{``ideal hyperparameters"}.

The aim of this experiment is to test the robustness of our approach w.r.t hyperparameter choices, i.e., how much our approach relies on labeled development data. Similar to the experiment in Section \ref{sec:Overall_evaluation}, we report the MCC scores on in-domain setting, i.e., WMT21 test data. Here we apply the hyperparameter settings in an opposite way compared to previous experiment, i.e., (1) applying the best hyperparameter setting of \textit{en-de} on \textit{en-zh} and \textit{en-ja}, and (2) applying the best hyperparameter setting of \textit{en-zh} on \textit{en-de} and \textit{en-cs}. We refer to this as \textit{``suboptimal hyperparameters"}. The MCC scores reducing significantly would indicate that our approach is sensitive to hyperparameters and vice versa. Additionally, we provide ablation experiments on the discrete hyperparameters to see their effects on the QE performance.

\section{Results and Discussion}
\subsection{Overall QE performance}

\begin{table}[htbp]

\centering
\begin{tabular}{lcc:cc}
\toprule
                     & \multicolumn{2}{c:}{Supervised/Tuned} & \multicolumn{2}{c}{Zero-shot/Un-tuned} \\
                     & en-de             & en-zh            & en-ja                  & en-cs         \\ \hline
Log probability QE & 0.241             & 0.149            & 0.112                  & 0.257         \\
WMT21 QE baseline & \textbf{0.370}    & \textbf{0.247}   & 0.131                  & 0.273         \\ \hline
Perturbation-based QE    & 0.287             & 0.180            & \textbf{0.218}         & 0.270        \\
\bottomrule
\end{tabular}

\caption{Performance (in MCC score) of word-level QE approaches on WMT21 test data. 
}
\label{tab:overal}
\end{table}

The performance of our Perturbation-based QE on the WMT21 test data, in comparison to other baselines, is shown in Table \ref{tab:overal}. Our approach outperforms the log-probability baseline over all language pairs. 
The largest gap is on \textit{en-ja}, where our approach obtains 0.106 points higher. 
For \textit{en-de} and \textit{en-zh}, the gain from our approach is around 0.040 points. 
The smallest gain is on \textit{en-cs}, where our approach outperforms the log-probability baseline by 0.013 points.

Compared to the WMT21 QE baseline, on \textit{en-de} and \textit{en-zh}, our approach falls behind by 0.083 and 0.067 points respectively. 
Recall that for these language pairs, the WMT21 QE is supervised, requiring labeled word-level data, and additionally uses sentence-level data for multi-task training. 
In contrast, our approach only uses the development split of word-level data for hyperparameter tuning. 
It can be concluded that our approach is not competitive with the supervised approaches that make use of more labeled data for training.

On \textit{en-ja} and \textit{en-cs}, our approach is competitive to the WMT21 QE baseline. 
Our approach outperforms the baseline by 0.087 points on \textit{en-ja}, while having similar performance on \textit{en-cs}. 
Recall that on these language pairs, both approaches do not use any labeled data of the language pairs in consideration. 
The WMT21 baseline is zero-shot: it uses labeled training data of 7 other language pairs. 
In contrast, our approach only uses the development split of 2 other language pairs for hyperparameter tuning. 
It can be concluded that, when there is no direct labeled data for the language pair of interest, our approach has competitive performance while being more data efficient compared to the zero-shot approach. 
Additionally, observe that the performance of the WMT21 zero-shot baseline on \textit{en-ja} is low compared to \textit{en-cs}. 
This is possibly due to the low similarity between \textit{ja} and other languages in the training data. 
This indicates that the zero-shot approach is more data-dependent, while this is not an issue for our approach.

\subsection{Out-of-domain, unseen-system evaluation}

The QE approaches' performance on detecting out-of-domain errors from known/unseen evaluated MT-system is shown in Table \ref{tab:OOD_error_detection}. As can be seen, our Perturbation-based QE approach has the best performance. 
When the task is to detect gender errors, our GenderBAD-recall is significantly higher than the second-best QE approach by over 0.200 points. When the task is to detect word sense disambiguation errors, our WSD-BAD-recall is higher than the second-best QE approach by over 0.049 points.
At the same time, our GenderBAD-precision and WSD-BAD-precision scores are similar to the other methods, indicating that we are not overly predicting tokens as BAD to cheat for a higher GenderBAD-recall and WSD-BAD-recall.

\begin{table}[h]
\centering
\begin{tabular}{lcccc}
\toprule
                      & \multicolumn{2}{c}{\begin{tabular}[c]{@{}c@{}}Known MT system\\ \underline{\hspace{0.3cm}(Encoder-Decoder MT)\hspace{0.3cm}}\end{tabular}}                                  & \multicolumn{2}{c}{\begin{tabular}[c]{@{}c@{}}Unseen MT system\\ \underline{\hspace{0.5cm}(Prompt-based LLM)\hspace{0.5cm}}\end{tabular}}                            \\
                      & \begin{tabular}[c]{@{}c@{}}GenderBAD\\ recall\end{tabular} & \begin{tabular}[c]{@{}c@{}}GenderBAD\\ precision\end{tabular} & \begin{tabular}[c]{@{}c@{}}GenderBAD\\ recall\end{tabular} & \begin{tabular}[c]{@{}c@{}}GenderBAD\\ precision\end{tabular} \\
Log probability QE    & 0.175                                                      & 0.036                                                          & 0.429                                                      & 0.047                                                          \\
WMT21 QE (supervised) & 0.021                                                      & 0.031                                                          & 0.065                                                      & 0.036                                                          \\
Perturbation-based QE & \textbf{0.391}                                             & 0.045                                                          & \textbf{0.658}                                             & 0.042                                                          \\
\hline
                      & \multicolumn{2}{c}{\begin{tabular}[c]{@{}c@{}}Known MT system\\ \underline{\hspace{0.3cm}(Encoder-Decoder MT)\hspace{0.3cm}}\end{tabular}}                                  & \multicolumn{2}{c}{\begin{tabular}[c]{@{}c@{}}Unseen MT system\\ \underline{\hspace{0.5cm}(Prompt-based LLM)\hspace{0.5cm}}\end{tabular}}                            \\
                      & \begin{tabular}[c]{@{}c@{}}WSD-BAD\\ recall\end{tabular}   & \begin{tabular}[c]{@{}c@{}}WSD-BAD\\ precision\end{tabular}   & \begin{tabular}[c]{@{}c@{}}WSD-BAD\\ recall\end{tabular}   & \begin{tabular}[c]{@{}c@{}}WSD-BAD\\ precision\end{tabular}   \\
Log probability QE    & 0.137                                                      & 0.007                                                          & 0.347                                                      & 0.005                                                          \\
WMT21 QE (supervised) & 0.290                                                      & 0.028                                                          & 0.177                                                      & 0.004                                                          \\
Perturbation-based QE & \textbf{0.339}                                             & 0.009                                                          & \textbf{0.709}                                             & 0.005                                                          \\
\bottomrule
\end{tabular}

\caption{Results on detecting out-of-domain errors by different QE methods. The top half indicates results on WinoMT. The bottom half indicates results on MuCoW. 
}
\label{tab:OOD_error_detection}

\end{table}

Another observation is that, the supervised WMT21 QE model performs poorly on detecting GenderBAD tokens from MT outputs on WinoMT. Its GenderBAD-recall is very low, at 0.021 for known MT system and at 0.065 for unseen MT system. 
This performance is even worse than that of the naive Log probability QE, whose GenderBAD-recall is 0.175 on known MT system output and 0.429 on unseen MT system output. 
A possible explanation is that, the WMT21 QE model does not aware of gender errors since this type of error does not present in the training data.
This indicates the data-dependent issue inherent to supervised QE approaches.

The performance gap between our Perturbation-based QE approach and the supervised WMT21 QE model is larger on the unseen MT system than the known MT system. 
On the unseen MT system, our GenderBAD-recall is higher than the WMT21 QE model by +0.593 points, which is larger than the corresponding gap of +0.370 when evaluating the known MT system. 
Similarly, the WSD-recall gap is +0.532 on the unseen MT system, which is larger than the gap of +0.049 on the known MT system. 
Additionally, when detecting word sense disambiguation errors from the unseen MT system, the performance of the WMT21 QE model is once again worse than the naive Log probability QE approach, where its WSD-BAD-recall is lower by -0.170 points. 
This indicates the MT-system-dependent issue inherent to supervised QE approaches, where they are not able to perform as well on evaluating unseen MT systems.

Overall, the supervised QE models, while performing better in a similar setting as their training process, fail behind Perturbation-based QE in out-of-domain and unseen-system settings. 
This strengthens the domain-independent and system-independent power of our approach: it can better detect errors from a new MT system on a new domain usage.

\subsection{Robustness w.r.t hyperparameter choices}
The difference in performance when using different hyperparameter settings in our approach is shown in Table \ref{tab:subop_hyp}. As can be seen, the MCC scores only deviate by around 0.010 points when using ideal versus suboptimal hyperparameter values.
The difference in performance when using different values for the discrete hyperparameters is shown in Table \ref{tab:diff_settings}. 
We consider 3 hyperparameters, in the top-to-bottom order displayed in Table \ref{tab:diff_settings}: sets of perturbed source words, unmasking models and alignment methods. 
It can be seen that, the MCC scores generally deviate by only around 0.010 points. 
Overall, different choices of hyperparameters do not significantly affect our QE performance. 
This shows that our approach is not sensitive to hyperparameter choices, which is useful since we are not dependent on labeled data for hyperparameter tuning.

\begin{table}[htbp]
\centering
\begin{tabular}{lllll}
\toprule
                                                 & \hspace{0.5cm}en-de                       & \hspace{0.5cm}en-zh                       & \hspace{0.5cm}en-ja                       & \hspace{0.5cm}en-cs                      \\ \hline
Ideal hyperparameters                            & \hspace{0.5cm}0.287                       & \hspace{0.5cm}0.180                       & \hspace{0.5cm}0.218                       & \hspace{0.5cm}0.270                      \\
Suboptimal hyperparameters                       & \hspace{0.5cm}0.274                       & \hspace{0.5cm}0.167                       & \hspace{0.5cm}0.206                       & \hspace{0.5cm}0.284                      \\ 
\bottomrule
\end{tabular}
\caption{Perturbation-based QE performance in MCC using ideal/suboptimal hyperparameters.}
\label{tab:subop_hyp}
\end{table}

\begin{table}[htbp]
\centering
\begin{tabular}{lllll}
\toprule
\multicolumn{1}{l}{}                       & \multicolumn{2}{c}{en-de}                               & \multicolumn{2}{c}{en-zh}                               \\
\multicolumn{1}{l}{}                       & \multicolumn{1}{c}{Best val MCC} & \multicolumn{1}{c}{Test MCC} & \multicolumn{1}{c}{Best val MCC} & \multicolumn{1}{c}{Test MCC} \\ \hline
\multicolumn{1}{l}{content words}          & 0.282 ± \small{\textcolor{gray}{0.004}}       & 0.287 ± \small{\textcolor{gray}{0.005}}   & 0.196 ± \small{\textcolor{gray}{0.003}}                & 0.169 ± \small{\textcolor{gray}{0.005}}            \\
\multicolumn{1}{l}{all words}              & 0.267 ± \small{\textcolor{gray}{0.004}}                & 0.277 ± \small{\textcolor{gray}{0.003}}            & 0.196 ± \small{\textcolor{gray}{0.002}}                & 0.162 ± \small{\textcolor{gray}{0.004}}            \\
\multicolumn{1}{l}{all tokens}             & 0.270 ± \small{\textcolor{gray}{0.002}}                & 0.269 ± \small{\textcolor{gray}{0.006}}            & 0.208 ± \small{\textcolor{gray}{0.003}}       & 0.177 ± \small{\textcolor{gray}{0.004}}   \\ \hline
\multicolumn{1}{l}{BERT large}       & 0.272 ± \small{\textcolor{gray}{0.006}}                & 0.278 ± \small{\textcolor{gray}{0.011}}            & 0.198 ± \small{\textcolor{gray}{0.006}}                & 0.174 ± \small{\textcolor{gray}{0.006}}            \\
\multicolumn{1}{l}{BERT base}        & 0.272 ± \small{\textcolor{gray}{0.006}}                & 0.279 ± \small{\textcolor{gray}{0.009}}            & 0.199 ± \small{\textcolor{gray}{0.005}}                & 0.166 ± \small{\textcolor{gray}{0.006}}            \\
\multicolumn{1}{l}{DistilBERT}  & 0.269 ± \small{\textcolor{gray}{0.008}}                & 0.272 ± \small{\textcolor{gray}{0.008}}            & 0.198 ± \small{\textcolor{gray}{0.008}}                & 0.168 ± \small{\textcolor{gray}{0.009}}            \\
\multicolumn{1}{l}{RoBERTa}           & 0.277 ± \small{\textcolor{gray}{0.007}}                & 0.278 ± \small{\textcolor{gray}{0.008}}            & 0.203 ± \small{\textcolor{gray}{0.007}}                & 0.169 ± \small{\textcolor{gray}{0.007}}            \\ \hline
\multicolumn{1}{l}{Levenshtein}            & 0.272 ± \small{\textcolor{gray}{0.006}}                & 0.275 ± \small{\textcolor{gray}{0.008}}            & 0.199 ± \small{\textcolor{gray}{0.006}}                & 0.170 ± \small{\textcolor{gray}{0.007}}            \\
\multicolumn{1}{l}{Tercom}                 & 0.274 ± \small{\textcolor{gray}{0.009}}       & 0.280 ± \small{\textcolor{gray}{0.009}}   & 0.200 ± \small{\textcolor{gray}{0.007}}                & 0.168 ± \small{\textcolor{gray}{0.007}}            \\ 
\bottomrule
\end{tabular}

\caption{Ablation experiments on discrete hyperparameter settings for Perturbation-based QE.
The performance of a setting in a specific group is averaged over all settings of the other groups.
}
\label{tab:diff_settings}

\end{table}

\subsection{Perturbation-based QE for explainable MT}

\begin{figure}[h]
  \centering
    \includegraphics[width=0.99\textwidth]{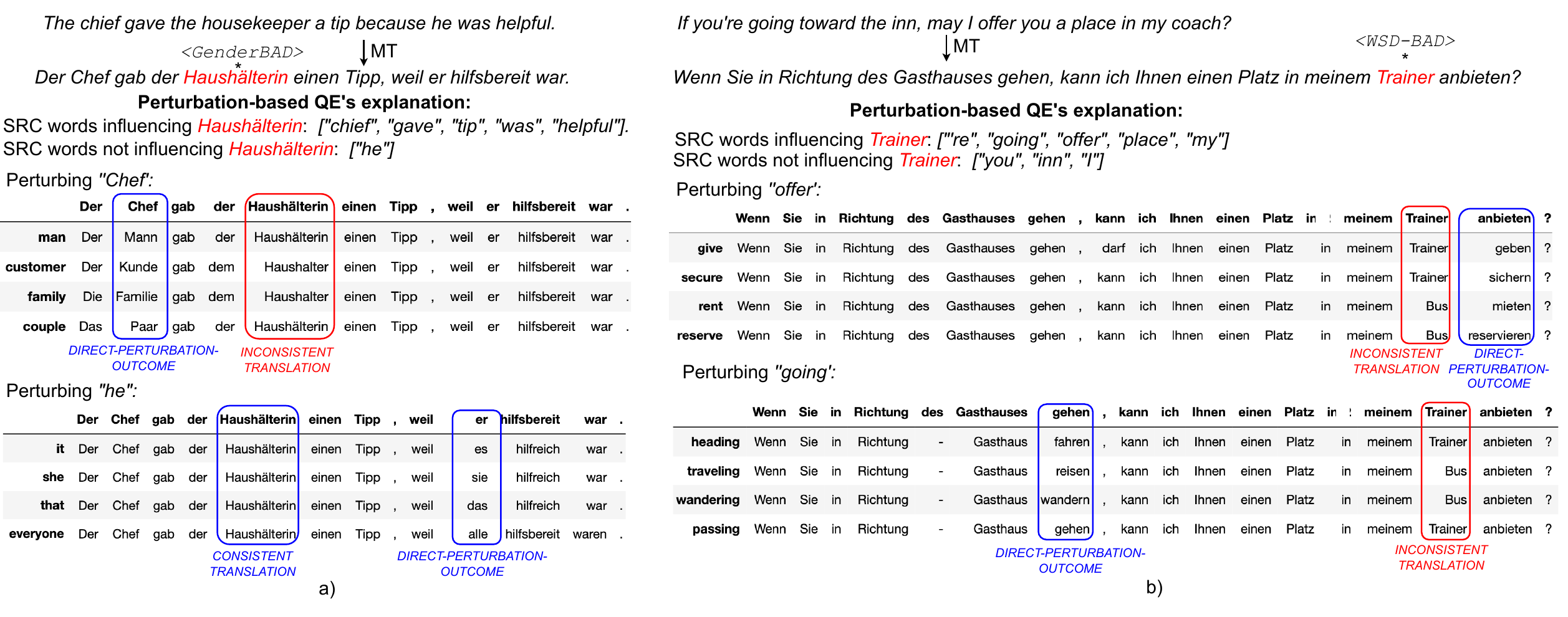}
    \caption{Example of Perturbation-based QE's explanation on WinoMT (a) and MuCoW (b).}
    \label{fig:ExplainabilityExample}
\end{figure}

We investigate some examples of using Perturbation-based QE for explaining the MT output gender errors on WinoMT and the word sense disambiguation errors on MuCoW. 
One gender error example is shown in Figure \ref{fig:ExplainabilityExample}a. 
The MT system outputs the female form for \textit{``housekeeper"}, while the form should be male, indicated by the word \textit{``he"}. 
In an ideal scenario, the gender form of \textit{``housekeeper"} (\textit{``Haushälterin/Haushalter"}) should only depend on \textit{``he"}. 
However, our approach shows that, when perturbing the source word \textit{``he"}, the output word \textit{``Haushälterin"} does not change. 
Instead, when perturbing \textit{[``chief", ``gave", ``tip", ``was", ``helpful"]}, the output varies between \textit{``Haushälterin"} and \textit{``Haushalter"}, showing that the MT model is focusing on the wrong part of the sentence to determine the gender form.

One word sense disambiguation error example is shown in Figure \ref{fig:ExplainabilityExample}b. The MT system outputs the wrong sense for \textit{``coach"}.  
It outputs \textit{``Trainer"}, which means the sports trainer. 
However, given the context, \textit{``coach"} should mean the vehicle, thus the correct output should be \textit{``Bus"}. 
Ideally, the MT system should only look at the context source words indicating movements to decide on the sense of \textit{``coach"}. 
Nevertheless, our approach shows that the MT output translation for \textit{``coach"} varies when perturbing multiple source words. 
For example, replacing \textit{``offer"} with \textit{``rent"} or \textit{``reserve"} makes the system outputs the correct sense \textit{``Bus"}, while for other replacements it still outputs \textit{``Trainer"}. 
Similarly, when replacing \textit{``going"} with other words that indicate movements, only \textit{``traveling"} and \textit{``wandering"} make the system outputs the correct sense \textit{``Bus"}. 
This explanation provides an insight into the MT system: the sense \textit{``Bus"} is only correlated to a few context words. 
Therefore, when the source sentence does not contain those specific words, the MT system fails to output the correct sense.

\section{Conclusion}
We proposed an unsupervised word-level Quality Estimation method, termed Perturbation-based QE. 
Our method does not rely on labeled QE data nor parallel MT data, masking it more domain-independent and system-independent to find MT errors that cannot be foreseen. 
This advantage is supported by our experiment on finding gender bias and word-sense-disambiguation erroneous translation from a nontraditional translation-prompting LLM.
Our approach is not sensitive to hyperparameter settings, thus less dependent on labeled data for hyperparameter tuning. 
Our approach is also explainable: it shows which source words affect an output translation word. 
Additionally, our approach is MT-system-agnostic and works for blackbox systems. 
Overall, Perturbation-based QE, as an unsupervised method, still falls behind supervised QE on in-domain and known-MT-system settings, but outperforms supervised QE on zero-shot settings and on out-domain and unseen-MT-system settings. As future work, it can be extended to
assess the quality of other tasks, such as question answering or summarization, in the same manner: minimally perturb the input and analyze changes in the output.

\small

\bibliographystyle{apalike}
\bibliography{anthology,custom}

\end{document}